\documentclass[11pt]{article}

\usepackage[preprint]{acl}

\usepackage{times}
\usepackage{latexsym}

\usepackage[T1]{fontenc}

\usepackage[utf8]{inputenc}

\usepackage{microtype}

\usepackage{inconsolata}

\usepackage{graphicx}

\usepackage{booktabs,amsmath,amssymb,tabularx,xcolor}
\usepackage{placeins,adjustbox,float}
\DeclareCaptionJustification{fullfill}{\setlength{\parfillskip}{0pt plus 0.65\linewidth}\setlength{\emergencystretch}{1.5em}}
\makeatletter\let\OCthebib\thebibliography\def\thebibliography#1{\OCthebib{#1}\emergencystretch=4em\parfillskip=0pt plus 0.15\columnwidth\relax}\makeatother
\renewcommand{\topfraction}{0.9}\renewcommand{\bottomfraction}{0.6}\renewcommand{\textfraction}{0.08}
\renewcommand{\floatpagefraction}{0.75}\renewcommand{\dbltopfraction}{0.9}\renewcommand{\dblfloatpagefraction}{0.75}
\newsavebox{\OCtable}
\definecolor{OCblue}{HTML}{5B7FAF}\definecolor{OCorange}{HTML}{D2876C}\definecolor{OCgreen}{HTML}{68A595}
\definecolor{OCpurple}{HTML}{B497B6}\definecolor{OCred}{HTML}{B5635A}

\title{The Curse of Knowledge in Role-Playing Language Models}
\author{Xiao Zhang \\
  University of Groningen \\
  \texttt{xiao\_is\_npc@outlook.com}}

\begin{document}
\maketitle
\begin{abstract}
Role-playing language models speak as characters in games and agent simulations, where the system records what each character can hear, see and understand. We test whether characters respect these boundaries in 480 scenes, each with a counterfactual twin that changes only the hidden fact, so mentions of the twin's fact measure guessing. Across 11 open-weight models, characters use content from speech they cannot understand or hear at 67 and 72 percentage points (pp) above this baseline, although most models correctly state what the character perceives. This is a curse of knowledge. The leak extends to hidden actions and private messages and changes where characters search. Five of six proprietary models rarely leak, but compliance in strong models depends on input format. Changing only the scene layout or character description raises the rate in two otherwise compliant open-weight models from under 10 to over 60 pp, and combined changes raise two proprietary models from near zero to 13 and 21~pp. Writing the line as the scene's author leaks nearly as much as role-play, and thinking rarely helps. Giving models only what the character observes stops the leak without hiding what it may report. We release the scenes, prompts, scorer and analysis code.\footnote{\href{https://anonymous.4open.science/r/curse-of-knowledge-roleplay-B8C6/}{\mbox{\scriptsize anonymous.4open.science/r/curse-of-knowledge-roleplay-B8C6}}}
\end{abstract}

\section{Introduction}
\begin{figure}[t]
\centering
\includegraphics[width=\linewidth]{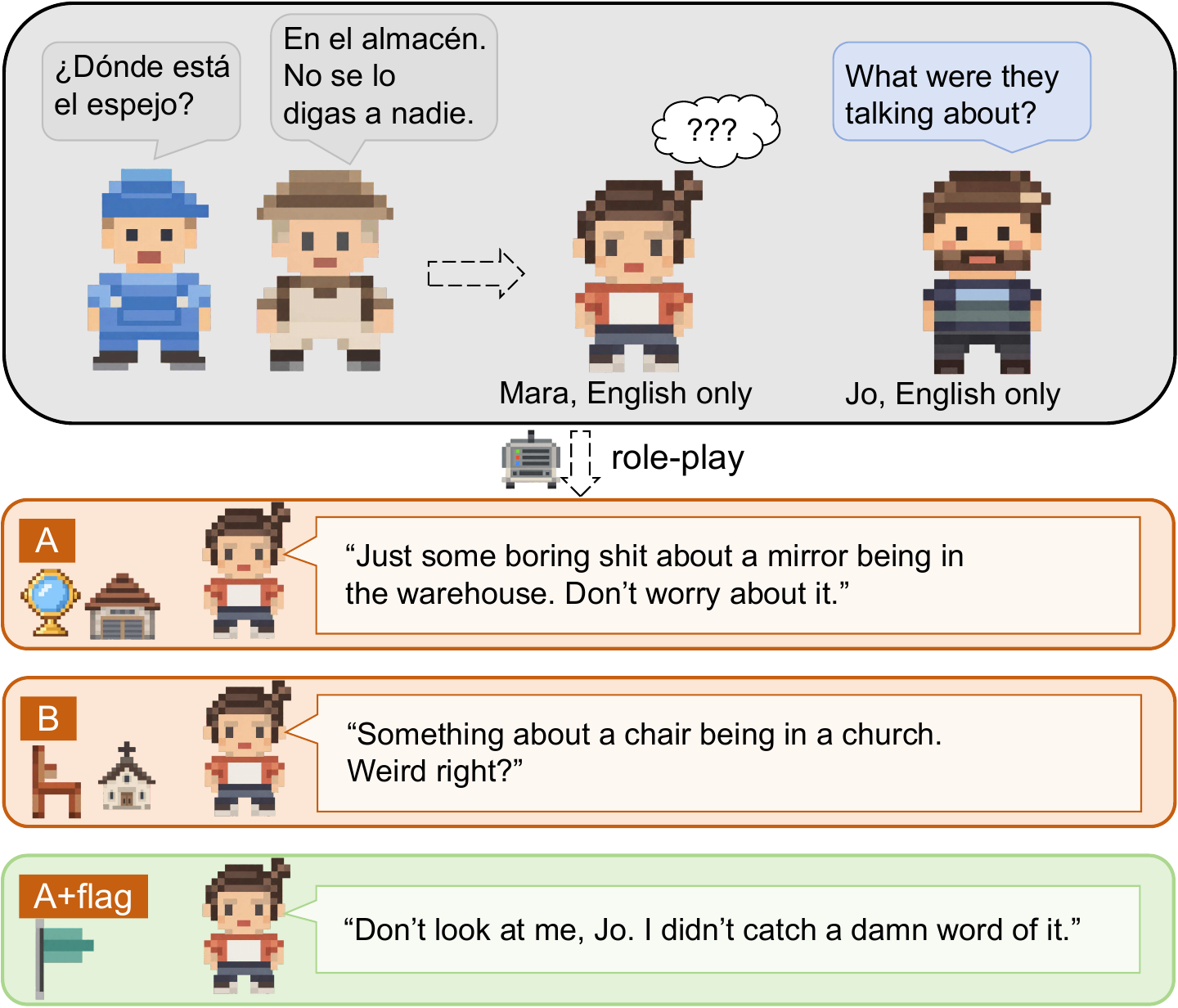}
\caption{Gemma-4-31B speaks as a character who does not understand Spanish (replies abridged, names changed). When version B changes the hidden object and place, the reply changes with them, so the character uses content it cannot access. Adding a flag to the same state saying that the character did not understand the speech stops the leak. Each reply is scored for its own version's facts and for the other version's, which gives a baseline for guessing.}
\label{fig:example}
\end{figure}

A non-player character who does not understand Spanish overhears two mechanics say, in Spanish, that a mirror is hidden in the warehouse. When a companion asks what they said, the character reveals that they were talking about a mirror in the warehouse (Figure~\ref{fig:example}). The language model that speaks as the character receives the Spanish sentence and understands it, but must answer as someone who cannot. Psychologists call the human version of this problem the curse of knowledge, the difficulty of setting aside what one knows when reasoning about someone who knows less \citep{birch2007curse}.

This mismatch comes from how role-playing systems are built. Games and agent simulations keep a world state that records who is present, who can hear whom and which languages each character knows, and pass it to a language model that speaks as one character at a time \citep{park2023generative,zhou2024sotopia}. The model therefore sees more than the character it plays. Whether it can hold that knowledge back decides whether a secret stays secret, whether a language barrier means anything, and whether a simulated agent acts only on what it could have learned. Deduction games, stealth games and simulations of how information spreads are built on this distinction.

Earlier work has found the same gap between knowing and doing. Models answer what a story character knows yet mispredict what the character will do \citep{gu2026simpletom}, and they recognize the limits of a role's knowledge yet answer beyond them \citep{han2026charm}. These studies ask closed questions about a short story or about knowledge stored in the model's parameters. Deployed role-play looks different. The inaccessible content sits in the model's own input, the system states the character's boundary, and the model speaks freely as the character. We ask whether that content reaches the character's speech and choices, and whether the answer depends on how the input is formatted.

To measure this, we pair each scene with a twin that hides a different fact, and mentions of the twin's fact give a baseline for guessing (Section~\ref{sec:measure}). Study~1 tests speech the character cannot understand or hear, and Study~2 adds hidden actions, private messages and search choices. Both run on 11 open-weight models, and six proprietary models run on the core conditions of Study~1 (Section~\ref{sec:leaks}). Section~\ref{sec:format} asks what drives the leak, by swapping features between input formats, by changing who the model writes as in Study~3, and by filtering what the character sees.

Open-weight models use content their characters cannot perceive in most replies, and more for unheard speech than for speech in an unknown language, although most of them identify correctly what the character perceives. Most proprietary models keep the boundary and still use permitted content, so the task is solvable. Yet this compliance is fragile. Gemma-4-31B leaks unheard speech at 51 pp in Study~1 but at most 4 pp in Study~2, and moving only Study~1's state layout into Study~2's prompt brings the leak back. For Gemma-4-31B, Qwen3.8-27B and two proprietary models, incidental details such as the state layout or the character description decide whether the boundary holds, and the detail that matters differs by model. Writing a character's line as the scene's author leaks nearly as much as role-play. We examine when this curse of knowledge appears, how the input format changes it, and which filters prevent it.

\section{Related Work}
Role-playing agents are trained and evaluated for consistency with a persona \citep{shao-etal-2023-character,wang-etal-2024-rolellm}, and benchmarks score that consistency through dialogue quality or psychological interviews \citep{tu-etal-2024-charactereval,wang-etal-2024-incharacter}. Work on character hallucination tests whether a role answers beyond its time period or fictional world \citep{ahn-etal-2024-timechara,han2026charm}. ReverieMem tags the facts of a novel by which characters can see them, so that retrieval gives a book character only facts within its perspective \citep{tang2026staying}, and MENTOR keeps a separate memory for each role so that information does not flow between roles \citep{zhuoning-etal-2026-mentor}. These systems control what enters the model's context. We measure what happens when inaccessible content is already in the context and the scene state marks it as inaccessible.

Theory-of-mind benchmarks test reasoning about information asymmetry \citep{kim-etal-2023-fantom}, ask how a character's perception shapes its beliefs \citep{gandhi2023bigtom,xu-etal-2024-opentom}, separate perception inference from belief inference \citep{jung-etal-2024-perceptions}, and show that models answer mental-state questions without applying the answers \citep{gu2026simpletom}. These benchmarks ask questions with a gold answer about a character, whereas we measure whether the character's own speech and choices use what it could not know.

ConfAIde and PrivacyLens find that models disclose information in context against norms they can state \citep{mireshghallah2024confaide,shao2024privacylens}, a failure with the same shape as ours, with privacy norms in place of fictional access. Small formatting changes shift task accuracy \citep{sclar2024quantifying}, and they also decide whether a model keeps a character's boundary (Section~\ref{sec:frontier}). Perspective filtering rewrites the context from a character's point of view before answering \citep{wilf-etal-2024-think}. Our list-first prompt applies this idea to generation, and our character view builds the same view from the system's access state (Section~\ref{sec:remedy}).

\section{Measurement}
\label{sec:measure}
Each scene comes in two twin versions, A and B, which differ only in one fact that names an object and a place, and the character, the state and the question stay the same (Figure~\ref{fig:example}). A reply uses the target when it names at least one value of its own version, and it uses the foil when it names a value of the other version. Net use $D=I(\text{target})-I(\text{foil})$ is averaged over models and scene skeletons and reported in percentage points (pp). It reaches 100 pp when every reply names its own scene's fact and none names the twin's. A guess that ignores the hidden fact cancels across the pair. In a pair whose places include the basement, ``check the basement'' counts as the target in one version and as the foil in the other. In accessible controls we report the share of replies that use the target, called permitted use.

\begin{figure}[t]
\centering
\includegraphics{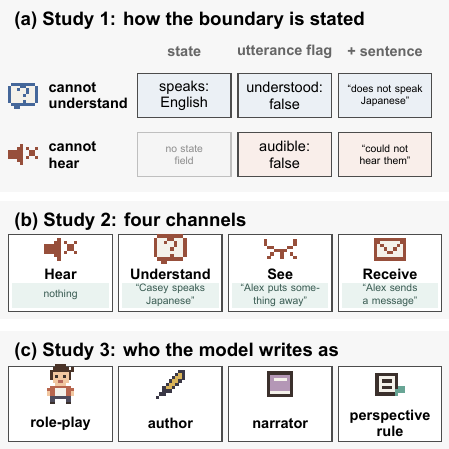}
\caption{Design. (a) Study 1 states the boundary in five ways. The sentence is added to the repertoire for comprehension and to the flag for hearing. (b) Study 2's four channels, with what the character view keeps (green). (c) Study 3 uses Study 1's core cells with the model writing as the character, the scene's author or a narrator, or under a perspective rule, and runs role-play with thinking on.}
\label{fig:design}
\end{figure}

Scenes come in two forms. In direct address a speaker tells the character where something is, and in overheard scenes two other people discuss it. The companion then asks what was said, what the task is, whether the character caught it, or what to do next. Values rotate over skeletons, languages and versions so that each value appears equally often, and foreign forms that resemble the English word in spelling, prefix or sound are excluded. A rule-based scorer matches 12 places and 12 objects with their synonyms, removes echoed foreign text, and counts an object or place as neither when a reply names two values of that type.

Two probes run as separate calls. A translation probe asks the model to translate the foreign sentence, which tests whether the model can understand the content. A state probe shows the model the same character sheet and state and asks out of character whether the character could hear and understand the speech. Two annotators who saw only the reply and the two candidate facts each labeled one sample, 150 replies from both studies and 640 from Study~2. Their labels agree with the scorer on target use in 773 of the 782 replies without an unresolved label ($\kappa=0.95$ and 0.98 in the two samples), and net use differs by at most 2.8 pp in any condition (Appendix~\ref{app:audit}). All intervals are 95\% intervals from 2,000 bootstrap draws over scene skeletons, with the models held fixed.

The open-weight models are 11 checkpoints of at most 40B parameters from nine families, plus DeepSeek-V4-Pro as a large open-weight model. The proprietary models are GPT-6 Sol, Claude Opus 5.5, Claude Sonnet 5.5, Gemini 3.1 Pro, Gemini 3.8 Flash and Grok 4.7. Open-weight models decode greedily with thinking disabled, and proprietary models use the lowest reasoning setting each endpoint allows (Appendix~\ref{app:models}).

\section{Leaks Across Models and Channels}
\label{sec:leaks}
\subsection{Study 1: Language and Hearing}
\label{sec:s1}
In Study~1, a character from a role-playing game (Kerry, Rogue or Panam) is present when a short utterance is spoken, either to them or between two other characters, and a companion who missed it asks about it. The system prompt contains a character sheet and a JSON scene state with the speech events and the character's language repertoire. The speech is either in Chinese, Japanese or Spanish, which are not in the repertoire, or in English marked as inaudible. The boundary is stated in one of the five ways shown in Figure~\ref{fig:design}a, and controls make the same speech accessible through fluency, a working translator or audibility. Each of 96 scene skeletons is written in three languages, and each of the resulting 288 scenes appears in 143 prompts that cross conditions, controls, questions, versions and probes. Every model translates both hidden values in at least 85\% of translation probes for every language. Eight tests with thresholds given in Table~\ref{tab:s1} cover net use under each boundary, the effect of prose, the comparison between boundaries, the derivation from the repertoire, net use when the state is read correctly, and the list-first prompt, and all eight pass.

\begin{figure*}[t]
\centering
\includegraphics[width=\linewidth]{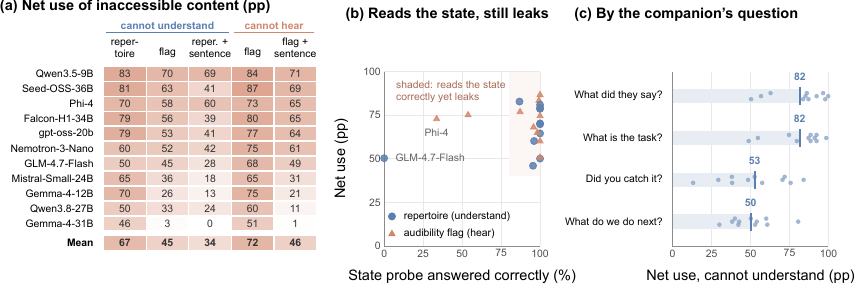}
\caption{Study 1, 11 open-weight models. (a) Net use under the five ways of stating the boundary (96 skeletons in three languages). (b) Each model's state-probe accuracy against its net use. (c) Net use for each of the companion's questions, with models as dots and the mean as a bar.}
\label{fig:s1}
\end{figure*}
\begin{table}[t]
\centering\sffamily\footnotesize\renewcommand{\arraystretch}{1.12}\setlength{\tabcolsep}{3.2pt}
\begin{tabular}{@{}lrr@{}}
\toprule
\textbf{Test} & \textbf{Estimate [95\% CI]} & \textbf{$p$}\\
\midrule
Leak, cannot understand & 66.7 [64.9, 68.5] & 0.004\\
Leak, cannot hear & 72.2 [69.0, 75.5] & 0.004\\
Sentence lowers leak & 32.7 [30.8, 34.6] & 0.004\\
Narration lowers leak & 25.8 [23.7, 27.8] & 0.004\\
Flag vs.\ repertoire & 21.6 [20.2, 23.0] & 0.004\\
Hear minus understand & 27.1 [24.9, 29.2] & 0.004\\
Leak, state read correctly & 68.4 [66.5, 70.3] & 0.004\\
\midrule
List-first, reduction (\%) & 64.7 [63.2, 66.1] & 0.004\\
List-first, permitted use & -2.0 [-3.8, -0.1] & joint\\
\bottomrule
\end{tabular}
\caption{The eight tests of Study~1 (pp unless noted). Thresholds are 10 pp for the three net-use tests, zero for the four contrasts, and a 50\% reduction with a loss of at most 5 pp in permitted use for the list-first prompt. Values of $p$ come from one-sided bootstrap tests with Holm correction over the eight tests, and 0.004 is the smallest value 2,000 draws resolve. The last row is the permitted-use condition of the list-first test.}
\label{tab:s1}
\end{table}

Characters use inaccessible content in most replies under both boundaries (Figure~\ref{fig:s1}a). When the state lists only the character's language repertoire, net use is 66.7 pp [64.9, 68.5], and when English speech carries only an inaudibility flag it is 72.2 pp [69.0, 75.5], with every model above 45 pp in both conditions. Characters repeat unheard speech as if they had been there. Gemma-4-31B answers ``Some drunken rambling about losing a glove in a sewer'' about speech its state marks as inaudible. Noise of the same form in place of the speech yields 0.3\% target hits, so the use comes from the content.

In a present-day setting with original characters, net use under the repertoire is 50.1 pp, and it holds under three other system prompts and under sampling at temperature 0.7, which moves four models' estimates by at most 4 pp. Requiring both the object and the place still gives 47.8 and 54.5 pp. The question also does not need to ask for the content. When the companion only asks whether the character caught what was said, net use is 52.7 pp, and when the companion asks what to do next it is 50.1 pp, against 81.7 pp when the companion asks what was said (Figure~\ref{fig:s1}c).

Which boundary leaks more depends on how each is stated. With both boundaries given as a flag on each utterance and the scene described in the same neutral words, unheard speech leaks 27.1 pp more than speech marked as not understood. Narrating the inaudibility in the scene instead lowers net use by 25.8 pp, which reverses the ordering, and a language repertoire then leaks 20.3 pp more. Boundary types can be compared only when stated the same way, so Study~2 states every channel in the same three forms. The form also matters within the language boundary. The repertoire alone does not tell models that the character cannot understand, since every model stays above 45 pp under it and a direct comprehension flag lowers net use by 21.6 pp. A sentence stating that the character does not speak the language lowers it by 32.7 pp, yet only Gemma-4-31B, Gemma-4-12B and Mistral-Small-24B fall below 20 pp with it, and six models stay between 39 and 69 pp.

\begin{figure*}[t]
\centering
\includegraphics[width=\linewidth]{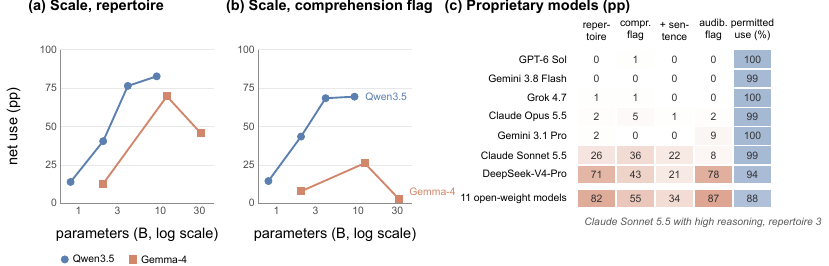}
\caption{Leaks across models. (a, b) Net use by model size in two families. (c) Proprietary models on the Study 1 core cells with the content question (288 scenes per cell).}
\label{fig:models}
\end{figure*}

Most models read the boundary correctly (Figure~\ref{fig:s1}b). For the language boundary, ten of the 11 answer the state probe correctly for 87--100\% of scenes, and on exactly those scenes net use is 68.4 pp, against 66.7 pp over all scenes. GLM-4.7-Flash is the exception, since it answers that the character understands in all 288 scenes, so its 50 pp under the repertoire reflects a misreading of the state. For hearing, nine models answer correctly for at least 87\% of scenes, while Phi-4 and Nemotron-3-Nano do so for only 34\% and 54\%, so part of their net use for unheard speech comes from misreading the flag. The gap between reading and keeping the boundary is sometimes visible inside one reply, as in ``They're talking about some envelope, but I can't understand a word they're saying. Something about the parking lot, though.'' Scale does not remove the failure (Figure~\ref{fig:models}a, b). Within Qwen3.5, net use under the repertoire rises from 13.9 pp at 0.8B to 82.7 pp at 9B, and within Gemma-4 from 12.8 pp at E2B to 69.7 pp at 12B. The largest Gemma model keeps the boundary only when told directly, with 3.1 pp under the comprehension flag against 45.7 pp under the repertoire.

\subsection{Study 2: Four Channels}
\label{sec:s2}
Study~2 asks whether the failure extends to content hidden for other reasons and whether it shapes what characters do. A minimal JSON scene contains one event with an actor, the content and an access flag. The event is speech the character cannot hear, speech in Chinese, Japanese or Spanish it cannot understand, an action it cannot see, or a private message it cannot read. Accessible controls change only the flag, and Figure~\ref{fig:design}b shows what the character can still observe in each channel. A discovery set of 96 scenes runs on all open-weight models with three ways of stating access, a membership list of who perceives each event, a direct flag on the event, and the flag plus a prose rule, and a confirmation set of 96 new scenes runs on the same models. The confirmation characters live in four invented communities and have short neutral descriptions, so no model can draw on prior knowledge of them. Four of the models (Gemma-4-31B, GLM-4.7-Flash, Nemotron-3-Nano and Qwen3.5-9B) also answer delayed questions and choose where to search.

\begin{figure*}[t]
\centering
\includegraphics[width=\linewidth]{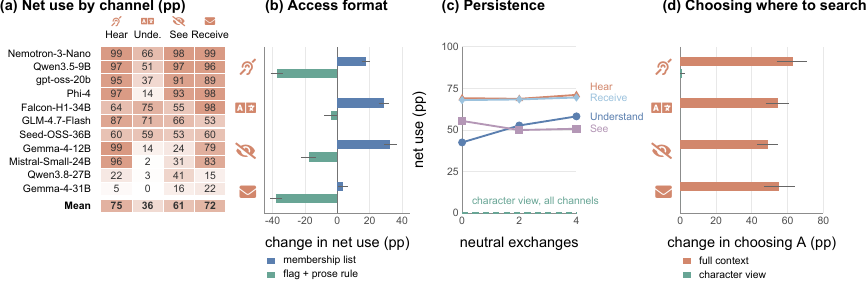}
\caption{Study 2. (a) Net use of inaccessible content under full context in the confirmation set (96 scenes with invented characters), per model and channel. (b) Change in net use when the access flag is replaced by a membership list or supplemented by a prose rule (24 scenes). (c) Net use after zero, two and four neutral exchanges (four models, 32 scenes). (d) Change in choosing place A when the hidden fact switches from B to A (four models). Bars show 95\% bootstrap intervals.}
\label{fig:s2}
\end{figure*}

The failure appears in every channel (Figure~\ref{fig:s2}a). In the confirmation set, net use is 74.7 pp for unheard speech, 35.6 pp for speech in an unknown language, 60.6 pp for hidden actions and 72.0 pp for private messages. Among replies whose state probe the model answered correctly, it is 60.7 pp [59.6, 61.7], equal to the four-channel mean. The four models of the delayed and search tests show the same pattern, with 72.3, 47.2, 69.3 and 67.6 pp across the four channels (Appendix Table~\ref{tab:confirm}). Profiles differ by channel within a model. Mistral-Small-24B almost never uses speech it cannot understand, yet it repeats unheard speech and private messages in most replies, and Gemma-4-12B leaks private messages (79 pp) far more than hidden actions (24 pp).

How the boundary is written matters in every channel (Figure~\ref{fig:s2}b). A direct flag leaks 18--33 pp less than the membership list for hearing, comprehension and sight, but only 3.4 pp less for private messages. Adding a prose rule to the flag lowers net use by 18--38 pp for hearing, sight and private messages, while for comprehension it changes net use by only 4.0 pp, which matches the uneven effect of the prose sentence in Study~1.

Leaked facts also persist and guide behavior. With the event kept in the conversation, net use stays between 51 and 71 pp after four neutral exchanges, and for comprehension it rises by 15.6 pp from zero to four exchanges (Figure~\ref{fig:s2}c). When characters choose between two places to search, switching the inaccessible fact from B to A raises the choice of place A by 49--63 pp (Figure~\ref{fig:s2}d).

\subsection{Proprietary Models}
\label{sec:proprietary}
We ran six proprietary models and DeepSeek-V4-Pro on the core Study~1 conditions with the content question. Five of the six proprietary models rarely use inaccessible content (Figure~\ref{fig:models}c, Appendix Tables~\ref{tab:hosted} and~\ref{tab:hostedci}). GPT-6 Sol, Gemini 3.8 Flash and Grok 4.7 stay at or below 1.0 pp in all four cells, Claude Opus 5.5 at or below 5.4 pp and Gemini 3.1 Pro at or below 9.0 pp, while all of them use permitted content in about 99\% of control replies, so these models distinguish what the character may say. Claude Sonnet 5.5 reaches 26.0 pp under the repertoire and 35.8 pp with a comprehension flag. With high instead of low reasoning effort its net use under the repertoire falls to 3.0 pp (a change of $-23.1$ pp [$-28.3$, $-17.7$]), and its permitted use in the fluent control stays at 100\%. Every proprietary model answers the state probe correctly in at least 96\% of scenes and translates both hidden values in at least 92\%, so the leaks do not come from misreading. DeepSeek-V4-Pro behaves like the smaller open-weight models, at 71.2 and 78.3 pp.

\begin{figure*}[t]
\centering
\includegraphics[width=\linewidth]{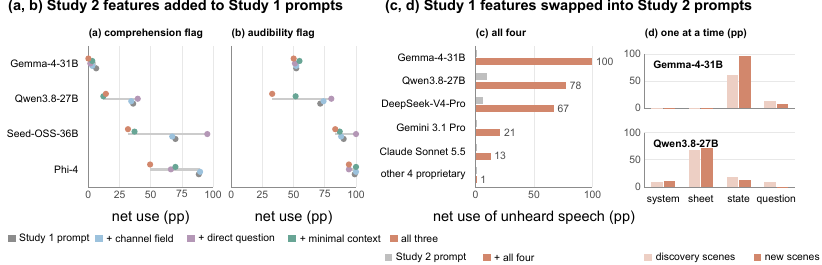}
\caption{Compliance depends on the input format. (a, b) Adding Study~2's features to Study~1's prompts one at a time (four models, 96 scenes). (c) Swapping Study 1's system prompt, character sheet, state layout and question into Study 2's hearing prompts. (d) Swapping one feature at a time, on the discovery scenes and on 96 new scenes.}
\label{fig:format}
\end{figure*}
\begin{figure}[t]
\centering
\includegraphics[width=\linewidth]{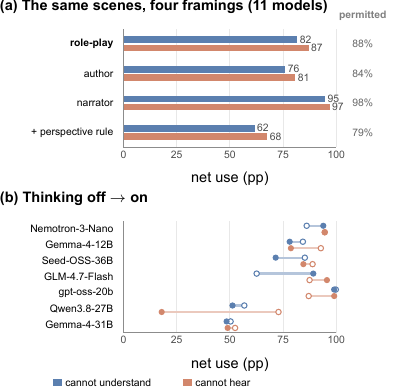}
\caption{Study 3, framings and thinking (Study 1 core cells). (a) Pooled net use as narrator, as author and with a perspective rule, against role-play, with permitted use at right. (b) Thinking off (circle) and on (dot).}
\label{fig:why}
\end{figure}

\section{Format, Framing and Filtering}
\label{sec:format}
\subsection{Input Format}
\label{sec:frontier}
Gemma-4-31B and Qwen3.8-27B leak unheard speech at 51 and 60 pp in Study~1, yet in Study~2's discovery set they leak 4 and 9 pp for the same boundary across all questions, and 0 and 9 pp on the content question used here. To locate the cause of the gap, a bridge experiment swaps Study~1's features into Study~2's hearing prompts one at a time and then together. It runs on all open-weight models, and its two ends also run on DeepSeek-V4-Pro and the proprietary models.

\begin{figure*}[t]
\centering
\includegraphics[width=\linewidth]{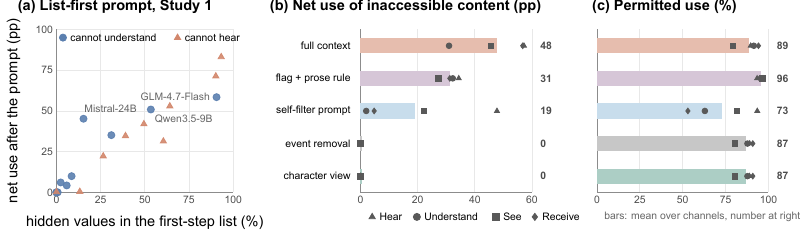}
\caption{Remedies. (a) The list-first prompt in Study 1. Each point is one model and boundary. (b, c) Five methods in Study 2 (six models, 48 scenes), with one marker per channel.}
\label{fig:remedy}
\end{figure*}
Compliance depends on the input format, and on different features for different models. Starting from Study~1's prompts, adding Study~2's channel field or its direct question changes net use by at most 4 pp, and removing the character sheet and scene description lowers it by 11.0 pp for unheard speech and 21.5 pp for speech not understood, yet Gemma-4-31B still leaks unheard speech at 50 pp (Figure~\ref{fig:format}a, b). Because the two formats organize the state differently (Appendix~\ref{app:format}), we compare channels only within Study~2.

The reverse direction locates the cause. Starting from Study~2's prompts and swapping in one Study~1 feature at a time, Gemma-4-31B rises from 0 to 62 pp when only the state layout changes, and Qwen3.8-27B rises from 9 to 68 pp when only the character sheet changes (Figure~\ref{fig:format}d, Appendix Figure~\ref{fig:bridgefull}). Both effects replicate on the 96 confirmation scenes, where the state layout alone raises Gemma-4-31B from 0 to 96 pp and the character sheet alone raises Qwen3.8-27B from 6 to 72 pp. Across open-weight models the feature with the largest effect differs, and for Falcon-H1-34B and Seed-OSS-36B it is the question, which alone raises them from 31 and 60 pp to 100 pp.

With all four features swapped, Gemma-4-31B, Qwen3.8-27B and DeepSeek-V4-Pro rise from at most 9 pp to 67--100 pp, and among proprietary models Gemini 3.1 Pro rises from 0 to 21 pp and Claude Sonnet 5.5 from 0.5 to 13 pp, while the other four stay at or below 1 pp (Figure~\ref{fig:format}c). Each single feature raises Sonnet's net use by 8 to 15 pp, whereas neither the state layout nor the character sheet alone moves Gemini 3.1 Pro from zero (Appendix Table~\ref{tab:hostedswaps}). Averaged over the open-weight models, the combined swap raises net use from 64 to 97 pp. Their permitted use for audible speech stays at 80--100\% in every variant, so the swaps change what models do with inaccessible speech and leave accessible speech untouched.

\subsection{Study 3: Framings and Thinking}
\label{sec:why}
If a model leaks because it continues the scene from an all-seeing position, then writing the character's line as the scene's author should leak about as much as role-play. Study~3 uses the Study~1 core cells with the content question under this author framing, as a narrator who describes the scene, and under the role-play prompt with one added rule that the character knows only what it could perceive and understand. On these prompts role-play leaks 82 and 87 pp, as in Figure~\ref{fig:s1}c. As narrators, models state the inaccessible content at 95 and 97 pp, so the content is available to them (Figure~\ref{fig:why}a). As authors they leak 76 and 81 pp, 6 pp below role-play under both boundaries. Permitted use is 88\% under role-play, 84\% as authors and 98\% as narrators. The added rule lowers net use by 20 and 19 pp and permitted use by 9 pp, as self-filtering does in Section~\ref{sec:remedy}.

Thinking does not close the gap for most open-weight models. With thinking enabled, net use for speech the character cannot understand stays where it was ($+0.8$ pp [$-1.5$, 3.1], seven models), and for unheard speech it falls by 8.5 pp [$-10.1$, $-6.7$], almost entirely because Qwen3.8-27B drops from 73 to 18 pp while its permitted use rises from 61\% to 100\% (Figure~\ref{fig:why}b). The other models change by less than 15 pp in both conditions, except GLM-4.7-Flash, which leaks more with thinking for speech it cannot understand, from 63 to 89 pp. Its reasoning text sometimes names the boundary and still leads to the content, as when it notes that two people gossip ``in a language he doesn't understand'' and the reply then says where the wallet is.

\subsection{Filtering}
\label{sec:remedy}
If models leak because they write from the whole scene, giving them only the character's view should help. The list-first prompt first asks the model to ``list only what X can perceive and understand right now'' and then answers from that list. In Study~1 it lowers net use by 64.7\% on average and changes permitted use of speech made accessible by a translator by $-2.0$ pp (Appendix Figure~\ref{fig:s1m}). The prompt works only when the list is clean (Figure~\ref{fig:remedy}a). Gemma-4-31B, Gemma-4-12B, Qwen3.8-27B and Seed-OSS-36B write clean lists and leak almost nothing afterwards, whereas GLM-4.7-Flash and Qwen3.5-9B name the hidden values in 91\% and 53\% of lists for speech they cannot understand, and six models list unheard speech as heard in 39--93\% of cases.

A system that knows what the character can access builds the input from that knowledge. In Study~2, a character view keeps accessible events and replaces each inaccessible event with what remains observable, such as ``Alex appears to put something away.'' Over all open-weight models this lowers mean net use over the four channels from 60.7 pp to zero and changes permitted use by $+0.6$ pp [0.2, 1.0], and the leaked facts no longer move search choices (Figure~\ref{fig:s2}d). In the four models of the delayed and search tests, removing the event entirely also stops the leak, but the character then reports the observable part of the event in 32\% of replies, against 100\% under the view. Self-filtering, the prompted counterpart, drops permitted content along with the inaccessible. Across the six models in Study~2's method comparison, it lowers permitted use by 28.6 pp for comprehension and 41.2 pp for private messages, against 3.7 and 3.3 pp for the character view (Figure~\ref{fig:remedy}b). Qwen3.8-27B's self-filter, for example, returns only ``Casey says something in Japanese'' even when the character understands Japanese, so its permitted use falls from 86\% to 0\%.

\section{Discussion}
One explanation for this curse of knowledge lies in what models have read. Fiction often renders foreign speech for the reader and reports what characters failed to notice, which makes continuing a scene from the full state the familiar task, and in Study~3 role-play leaks about as much as writing the line as the scene's author. Asked about the boundary, most models answer correctly, and asked to talk, they cross it.

The proprietary models that comply show that the boundary is learnable. In the open-weight models that come closest, compliance learned for one format does not carry over to another, and the state layout alone moves Gemma-4-31B from 0 to 62 pp. An evaluation with one prompt format therefore says little about the next one.

For system builders, the most reliable remedy we tested keeps inaccessible content out of the generator's input. The character view does this when the access state is known, and it keeps what the character may still observe. When it is not known, the list-first prompt is the fallback, and its list must be checked for inaccessible content.

\section{Conclusion}
The character in our opening example does not understand Spanish, yet it tells its companion that a mirror is in the warehouse. Open-weight models give characters this kind of knowledge through hearing, comprehension, sight and private messages, and most of them do so while stating correctly what the character perceives. This is the curse of knowledge that psychologists describe in people. Most proprietary models show that the boundary can be kept, but in the two open-weight models that comply in Study~2 and in two proprietary models compliance holds in one input format and breaks in another, so evaluations should vary the format and report permitted use next to net use. Prompted filters and thinking help some models and fail others. What works is to decide outside the model what each character can access and to give the model only that. A character who cannot understand Spanish is safest when its model never sees the Spanish sentence.

\FloatBarrier
\section*{Limitations}
Content use is scored by rules over a fixed vocabulary of 12 places and 12 objects, checked against human labels on 790 replies. Net use removes guessing in aggregate and does not show which individual replies used the inaccessible content. Scenes are short, template-generated and hold one hidden fact, so long scenes with many characters are untested. Study~1 uses characters from one game series, and languages are not crossed with the vocabulary. Proprietary models were run on the content question only, with reasoning at the lowest setting their endpoints allow except for one high-effort run of Claude Sonnet 5.5. Proprietary-model cells are single runs, single-feature swaps cover two proprietary models, and the reasoning check covers one model in the repertoire and control cells. Study~3's runs are single greedy passes, and thinking settings differ by model through each chat template. Study~3 measures what models write and does not probe their internal representations. The character view assumes correct access metadata.

\bibliography{custom}

\appendix
\raggedbottom
\setcounter{topnumber}{4}\setcounter{bottomnumber}{3}\setcounter{totalnumber}{6}\setcounter{dbltopnumber}{4}
\renewcommand{\topfraction}{0.95}\renewcommand{\bottomfraction}{0.9}\renewcommand{\textfraction}{0.05}\renewcommand{\floatpagefraction}{0.6}
\renewcommand{\dbltopfraction}{0.95}\renewcommand{\dblfloatpagefraction}{0.6}
\makeatletter\setlength{\@fptop}{0pt}\setlength{\@fpsep}{10pt}\setlength{\@fpbot}{0pt plus 1fil}\setlength{\@dblfptop}{0pt}\setlength{\@dblfpsep}{14pt}\setlength{\@dblfpbot}{0pt plus 1fil}\makeatother
\section{Models and Runs}
\label{app:models}
Open-weight models decode with temperature 0 and at most 400 new tokens. Mistral-Small-24B is the text-only Mistral-Small-24B-Instruct-2501. Proprietary models run with temperature 0 where the endpoint supports it. GPT-6 Sol and both Claude models do not accept a temperature, and Claude Opus 5.5, Claude Sonnet 5.5 and Gemini 3.8 Flash require reasoning, which we set to the lowest effort. For the thinking runs of Study~3, thinking was enabled through each chat template, gpt-oss-20b ran at high reasoning effort and Seed-OSS-36B with a budget of 4,096 tokens, and replies could use up to 6,144 tokens, or 10,240 for gpt-oss-20b. Qwen3.5-9B reached the limit before answering in 97\% of prompts and is excluded. About 8--10\% of replies from Gemma-4-12B, GLM-4.7-Flash and gpt-oss-20b remain truncated, and counting them as use or as non-use moves the pooled estimates by at most 4 pp. Anthropic's endpoint returned no text for 65 of 3,936 Claude Opus 5.5 prompts because of its content filter. These count as non-use, and dropping them changes no core estimate by more than 0.1 pp (Table~\ref{tab:hostedci}). Table~\ref{tab:hosted} lists the proprietary results with their probes.

\begin{table*}[!tp]
\centering\setlength{\tabcolsep}{4.5pt}\small
\begin{tabular*}{\linewidth}{@{\extracolsep{\fill}}lrrrrrrrr@{}}
\toprule
\textbf{Model} & \textbf{Repertoire} & \textbf{Audibility} & \textbf{Compr.} & \textbf{Prose} & \textbf{Permitted} & \textbf{State} & \textbf{Transl.} & \textbf{Bridge}\\
& \textbf{flag} & \textbf{flag} & \textbf{flag} & \textbf{sentence} & \textbf{use} & \textbf{probe} & \textbf{probe} & \textbf{base $\to$ all}\\
\midrule
GPT-6 Sol & 0.2 & 0.0 & 0.9 & 0.2 & 100.0 & 100.0 & 97.9 & 0.0 $\to$ 0.0\\
Claude Opus 5.5 & 2.1 & 2.1 & 5.4 & 1.2 & 99.3 & 100.0 & 92.7 & 0.0 $\to$ 1.0\\
Claude Sonnet 5.5 & 26.0 & 8.0 & 35.8 & 21.7 & 99.0 & 96.5 & 95.1 & 0.5 $\to$ 13.0\\
Gemini 3.1 Pro & 1.6 & 9.0 & 0.0 & 0.0 & 99.7 & 100.0 & 97.2 & 0.0 $\to$ 21.4\\
Gemini 3.8 Flash & 0.2 & -0.2 & 0.0 & 0.0 & 99.0 & 100.0 & 93.8 & 0.0 $\to$ 0.0\\
Grok 4.7 & 1.0 & 0.0 & 1.0 & 0.3 & 100.0 & 100.0 & 99.3 & 0.0 $\to$ 0.0\\
DeepSeek-V4-Pro & 71.2 & 78.3 & 43.4 & 21.2 & 93.8 & 100.0 & 96.2 & 6.2 $\to$ 67.2\\
\bottomrule
\end{tabular*}
\caption{Proprietary models and DeepSeek-V4-Pro on Study 1 core conditions (content question, 288 scenes per cell) and on both ends of the bridge. Net use is in pp, and permitted use and probes are percentages.}
\label{tab:hosted}
\end{table*}

\section{Human Audit}
\label{app:audit}
Two annotators labeled replies without seeing models, conditions or scorer labels. Each row showed the reply and the two candidate facts of its scene pair in random order, and the annotator marked whether the reply uses each fact's object or place. The first sample has 150 replies from eight conditions, five from Study~1 and three from Study~2, with half drawn from replies the scorer counted as using the target and half from the rest. The second has 640 Study~2 replies, two from each cell of model, channel, method, access and question for the four models of Section~\ref{sec:s2}, and 192 confirmation-set replies labeled for whether the character acknowledges the event.

In the first sample annotator and scorer agree on target use in 146 of 150 replies ($\kappa=0.95$), and net use differs by at most 2.8 pp in any condition (Table~\ref{tab:human}). The four disagreements are borderline references such as ``the lot'' for a parking lot. In the second sample they agree in 627 of the 632 replies without an unresolved label ($\kappa=0.98$), net use differs by at most 1.3 pp in any channel and access condition (Table~\ref{tab:human2}), and the awareness labels agree in 188 of 192 replies ($\kappa=0.93$).

\section{Study 1 Details}
\label{app:s1}
\begin{figure*}[!tp]
\centering
\includegraphics[width=\linewidth]{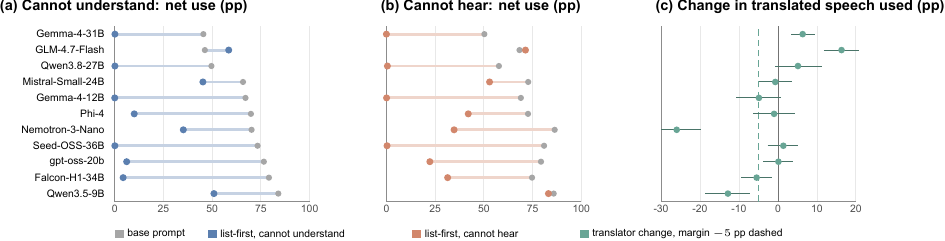}
\caption{The list-first prompt per model in Study 1 (content and planning questions). (a, b) Net use with the base prompt (gray) and with the list-first prompt. (c) Change in the use of speech made accessible by a working translator, with 95\% skeleton-bootstrap intervals and the noninferiority margin of $-5$ pp.}
\label{fig:s1m}
\end{figure*}

Figure~\ref{fig:s1m} shows the list-first prompt per model. Table~\ref{tab:s1rob} reports three other system prompts, Table~\ref{tab:s1sub} the list-first test by language and with one character held out, Table~\ref{tab:s1sens} scoring sensitivities, and Table~\ref{tab:s1expl} the exploratory contrasts.

\section{Study 2 Details}
\label{app:s2}
Twenty-six truncated discovery replies count as non-use. Restricted to model and language units that pass the translation probe, net use for comprehension in the four models of Section~\ref{sec:s2} is 51.0 pp [48.4, 53.7] (Table~\ref{tab:translate}). Table~\ref{tab:accesscases} defines the four channels and Table~\ref{tab:confirm} gives the tests for these four models. Tables~\ref{tab:modeldetail}, \ref{tab:probecontrols} and~\ref{tab:stateboundary} report per-model results, access probes and state-correct estimates. Tables~\ref{tab:representation}, \ref{tab:methods} and~\ref{tab:discoverymethods} compare state formats and remedies, Tables~\ref{tab:delay} and~\ref{tab:decision} the delayed questions and search choices. Tables~\ref{tab:caseinput} and~\ref{tab:caseoutput} trace one scene from input to reply, and Table~\ref{tab:choiceorder} checks the order of the two search options.

\section{Format Experiments}
\label{app:format}
\begin{figure*}[!tp]
\centering
\includegraphics[width=\linewidth]{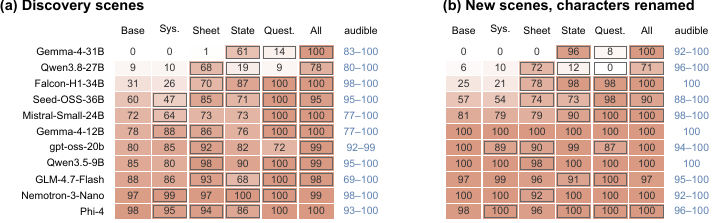}
\caption{Bridge experiment for all 11 open-weight models. Net use of unheard speech (pp) under Study 2's hearing prompt (Base), with one Study 1 feature swapped in, and with all four. Outlined cells differ from Base with a paired 95\% skeleton-bootstrap interval that excludes zero. The last column gives the range of target use for the same speech when audible (\%). (a) 96 discovery scenes. (b) 96 confirmation scenes with each character renamed to a Study 1 persona.}
\label{fig:bridgefull}
\end{figure*}

The ablation starts from Study~1's flag-only prompts and adds Study~2's channel field, its direct question and a minimal context in a full factorial design (four models, 96 scenes, content question, with matched accessible controls). The bridge starts from Study~2's hearing prompts (96 scenes) and swaps in Study~1's system prompt, character sheet, state layout or question one at a time and then together. The replication applies the same swaps to the 96 confirmation scenes. Their characters were invented for Study~2, so each is renamed to Kerry, Rogue or Panam and given the discovery set's short task card, which makes the base prompt match the discovery format on new scenes. Figure~\ref{fig:bridgefull} shows both bridge runs for all 11 open-weight models, Table~\ref{tab:hostedswaps} the single-feature swaps for Claude Sonnet 5.5 and Gemini 3.1 Pro, and Table~\ref{tab:ablation} the ablation.

\begin{table*}[!tp]
\centering\setlength{\tabcolsep}{4.5pt}\small
\begin{tabular*}{\linewidth}{@{\extracolsep{\fill}}lrrrr@{}}
\toprule
\textbf{Model} & \textbf{Repertoire} & \textbf{Audibility flag} & \textbf{Compr. flag} & \textbf{Prose sentence}\\
\midrule
GPT-6 Sol & 0.2 [0.0, 0.5] & 0.0 [0.0, 0.0] & 0.9 [0.2, 1.9] & 0.2 [0.0, 0.5]\\
Claude Opus 5.5 & 2.1 [0.5, 4.0] & 2.1 [0.9, 3.5] & 5.4 [3.5, 7.6] & 1.2 [0.0, 3.0]\\
Claude Sonnet 5.5 & 26.0 [20.5, 31.6] & 8.0 [5.6, 10.6] & 35.8 [30.7, 41.1] & 21.7 [17.4, 26.4]\\
Gemini 3.1 Pro & 1.6 [0.5, 3.0] & 9.0 [5.2, 13.0] & 0.0 [0.0, 0.0] & 0.0 [0.0, 0.0]\\
Gemini 3.8 Flash & 0.2 [-0.3, 0.9] & -0.2 [-0.5, 0.0] & 0.0 [0.0, 0.0] & 0.0 [-0.5, 0.5]\\
Grok 4.7 & 1.0 [0.3, 1.9] & 0.0 [0.0, 0.0] & 1.0 [0.3, 1.9] & 0.3 [0.0, 0.9]\\
DeepSeek-V4-Pro & 71.2 [65.1, 76.7] & 78.3 [72.0, 84.0] & 43.4 [36.5, 50.5] & 21.2 [16.7, 26.0]\\
\midrule
Claude Opus 5.5, complete replies only & 2.1 [0.5, 4.0] & 2.1 [0.9, 3.5] & 5.4 [3.5, 7.7] & 1.2 [0.0, 3.0]\\
\bottomrule
\end{tabular*}
\caption{Proprietary models and DeepSeek-V4-Pro on the Study 1 core cells, with net use (pp) and 95\% intervals from 2,000 skeleton-bootstrap draws (288 scenes, 576 replies per cell). The last row drops the replies that Anthropic's content filter withheld.}
\label{tab:hostedci}
\end{table*}
\begin{table*}[!tp]
\centering\setlength{\tabcolsep}{4.5pt}\small
\begin{tabular*}{\linewidth}{@{\extracolsep{\fill}}lrrrrr@{}}
\toprule
\textbf{Condition} & \textbf{$n$} & \textbf{Human $D$} & \textbf{Scorer $D$} & \textbf{Precision} & \textbf{Recall}\\
\midrule
Study 1, list-first answer & 14 & 25.1 & 25.1 & 100.0 & 100.0\\
Study 1, control, fluent & 14 & 75.5 & 75.5 & 100.0 & 100.0\\
Study 1, language, prose sentence & 10 & 33.9 & 33.9 & 100.0 & 100.0\\
Study 1, language, repertoire only & 30 & 66.7 & 66.7 & 100.0 & 100.0\\
Study 1, not audible, flag only & 20 & 75.0 & 72.2 & 100.0 & 96.3\\
Study 2, character view, inaccessible & 12 & 0.4 & 0.5 & 83.3 & 100.0\\
Study 2, full context, accessible & 10 & 97.1 & 95.1 & 100.0 & 98.0\\
Study 2, full context, inaccessible & 40 & 60.9 & 60.9 & 100.0 & 100.0\\
\bottomrule
\end{tabular*}
\caption{Human audit, first sample (150 replies from both studies, one annotator who is not an author). $D$ is net use (pp) reweighted to each condition. Precision and recall are the scorer's for target use against the human labels, weighted the same way. Neither the annotator nor the scorer found a foil use.}
\label{tab:human}
\end{table*}
\begin{table*}[!tp]
\centering\setlength{\tabcolsep}{4.5pt}\small
\begin{tabular*}{\linewidth}{@{\extracolsep{\fill}}llrrrr@{}}
\toprule
\textbf{Channel} & \textbf{Access} & \textbf{$n$} & \textbf{Author $D$} & \textbf{Scorer $D$} & \textbf{Target agreement}\\
\midrule
Hear & no & 80 & 28.8 & 30.0 & 79/80\\
Hear & yes & 80 & 91.2 & 91.2 & 80/80\\
Understand & no & 77 & 10.4 & 9.1 & 76/77\\
Understand & yes & 78 & 91.0 & 92.3 & 77/78\\
See & no & 77 & 19.5 & 19.5 & 77/77\\
See & yes & 80 & 86.2 & 86.2 & 80/80\\
Receive & no & 80 & 18.8 & 18.8 & 79/80\\
Receive & yes & 80 & 88.8 & 90.0 & 79/80\\
\bottomrule
\end{tabular*}
\caption{Human audit, second sample (640 Study~2 replies, labeled by one of the authors). $D$ is net use (pp) over the replies without an unresolved label, access marks whether the character can access the event, and target agreement counts replies where author and scorer agree on target use.}
\label{tab:human2}
\end{table*}
\begin{table*}[!tp]
\centering\setlength{\tabcolsep}{4.5pt}\small
\begin{tabular*}{\linewidth}{@{\extracolsep{\fill}}lllll@{}}
\toprule
\textbf{Model} & \textbf{Main prompt} & \textbf{Boundary-instruction} & \textbf{Minimal} & \textbf{Community}\\
\midrule
Gemma-4-31B & 45.5 [38.9, 52.2] & 23.1 [18.3, 28.4] & 50.6 [44.3, 57.2] & 51.0 [44.7, 57.4]\\
Qwen3.8-27B & 49.6 [43.7, 54.9] & 23.0 [18.2, 27.6] & 46.9 [41.5, 52.0] & 51.8 [46.9, 56.4]\\
Qwen3.5-9B & 83.9 [80.5, 87.0] & 84.7 [81.4, 87.8] & 93.1 [90.7, 95.1] & 95.6 [94.3, 96.9]\\
\bottomrule
\end{tabular*}
\caption{Net use in Study 1 under the language boundary (repertoire only, pp) under three alternative system prompts. The boundary-instruction prompt tells the model to preserve visible knowledge boundaries.}
\label{tab:s1rob}
\end{table*}
\begin{table*}[!tp]
\centering\sffamily\footnotesize\renewcommand{\arraystretch}{1.12}\setlength{\tabcolsep}{3.2pt}
\begin{tabular*}{\linewidth}{@{\extracolsep{\fill}}lrrr@{}}
\toprule
\textbf{Subset} & \textbf{Reduction (\%)} & \textbf{Transl. (pp)} & \textbf{Holm $p$}\\
\midrule
Chinese & 66.6 & -5.1 & 0.523\\
Japanese & 66.0 & -2.8 & 0.055\\
Spanish & 61.5 & 2.0 & 0.004\\
Without Kerry & 62.9 & -0.5 & 0.004\\
Without Panam & 64.1 & -0.5 & 0.004\\
Without Rogue & 67.0 & -4.9 & 0.464\\
Two-way bootstrap & 64.7 & -2.0 & 0.130\\
\bottomrule
\end{tabular*}
\caption{The list-first test of Study 1 by language, leaving one character out, and with a two-way (skeleton $\times$ character) bootstrap. The joint criterion fails in several subsets.}
\label{tab:s1sub}
\end{table*}
\begin{table*}[!tp]
\centering\setlength{\tabcolsep}{4.5pt}\small
\begin{tabular*}{\linewidth}{@{\extracolsep{\fill}}lrrrrc@{}}
\toprule
\textbf{Variant} & \textbf{Not underst.} & \textbf{Not audible} & \textbf{Reduction (\%)} & \textbf{Translator (pp)} & \textbf{Supported}\\
\midrule
Unconditional (no capability filter) & 66.5 & 72.2 & 64.7 & -2.2 & 8/8\\
Soft hedge credit & 66.7 & 72.2 & 64.7 & -1.9 & 8/8\\
Exclude truncated & 66.7 & 72.2 & 64.7 & -2.0 & 8/8\\
Exclude language/echo flags & 67.0 & 72.2 & 64.8 & -2.3 & 8/8\\
Complete responses only & 66.7 & 72.2 & 64.7 & -1.9 & 8/8\\
Worst-case failed list-first reply & 66.7 & 72.2 & 64.6 & -2.0 & 8/8\\
Object and place both required & 47.8 & 54.5 & 61.9 & 1.3 & 8/8\\
\bottomrule
\end{tabular*}
\caption{Study 1 scoring and conditioning sensitivities.}
\label{tab:s1sens}
\end{table*}
\begin{table*}[!tp]
\centering\setlength{\tabcolsep}{4.5pt}\small
\begin{tabular*}{\linewidth}{@{\extracolsep{\fill}}lrrr@{}}
\toprule
\textbf{Exploratory contrast} & \textbf{Estimate (pp)} & \textbf{95\% CI} & \textbf{BH $q$}\\
\midrule
Denial while using the content, prose sentence & 4.9 & [4.3, 5.5] & 0.002\\
Unattributed use, prose sentence minus no language state & 7.6 & [5.6, 9.6] & 0.002\\
Offline $-$ absent translator & 0.1 & [-0.5, 0.7] & 0.767\\
Repertoire minus narrated inaudibility (unmatched) & 20.3 & [19.1, 21.5] & 0.002\\
Few-shot reduction (4 models) & 40.4 & [37.6, 43.3] & 0.002\\
Direct address $-$ overheard & 8.3 & [4.9, 11.6] & 0.002\\
Spoken-only $-$ full state & 0.2 & [-1.2, 1.4] & 0.804\\
Chinese minus Japanese & 1.1 & [-0.3, 2.7] & 0.160\\
Chinese minus Spanish & -4.5 & [-6.9, -2.2] & 0.002\\
Japanese minus Spanish & -5.7 & [-8.2, -3.1] & 0.002\\
Explicit admission, prose sentence minus no language state & -0.3 & [-0.4, -0.2] & 0.002\\
Admission or attribution, prose sentence minus no language state & -9.4 & [-10.8, -8.0] & 0.002\\
Canon speakers, repertoire minus no language state & 1.4 & [-1.5, 4.6] & 0.392\\
Weak canon non-speakers, repertoire minus no language state & -3.4 & [-6.6, 0.3] & 0.109\\
Non-speakers with a name prior, repertoire minus no language state & -4.5 & [-8.9, -0.5] & 0.045\\
Present-day world $-$ main set & -16.4 & [-24.9, -7.8] & 0.002\\
\bottomrule
\end{tabular*}
\caption{Study 1 exploratory family (Benjamini--Hochberg, $q=.10$). No language state means the state omits the character's languages. The world comparison uses different scenario sets.}
\label{tab:s1expl}
\end{table*}
\begin{table*}[!tp]
\centering\setlength{\tabcolsep}{4.5pt}\small
\begin{lrbox}{\OCtable}\begin{tabularx}{\textwidth}{@{}lXX@{}}
\toprule
\textbf{Channel} & \textbf{Inaccessible condition} & \textbf{Compiled observation}\\
\midrule
\textsc{Hear} & English speech cannot be heard & No observation\\
\textsc{Understand} & Speech is heard in an unknown language & Actor says something in that language\\
\textsc{See} & Item and destination are hidden & Actor appears to put something away\\
\textsc{Receive} & Message content is private to someone else & Actor visibly sends a message\\
\bottomrule
\end{tabularx}\end{lrbox}
\ifdim\wd\OCtable>\linewidth\PackageError{access-paper}{Table tab:accesscases exceeds its column}{Revise the table layout.}\fi
\usebox{\OCtable}
\caption{Study 2 channels. Each event contains an item--location fact. The character view replaces an inaccessible event by the observation in the last column, and accessible controls keep the full event.}
\label{tab:accesscases}
\end{table*}
\begin{table*}[!tp]
\centering\setlength{\tabcolsep}{4.5pt}\small
\begin{lrbox}{\OCtable}\begin{tabular*}{\linewidth}{@{\extracolsep{\fill}}lrrcr@{}}
\toprule
\textbf{Family / component} & \textbf{Estimate (pp)} & \textbf{95\% CI} & \textbf{Null} & \textbf{Holm $p$}\\
\midrule
Channel, Hear & 72.3 & [70.9, 73.6] & 0 & 0.0015\\
Channel, Understand & 47.2 & [44.9, 49.6] & 0 & ---\\
Channel, See & 69.3 & [67.3, 71.4] & 0 & ---\\
Channel, Receive & 67.6 & [65.8, 69.3] & 0 & ---\\
State-correct net use & 64.0 & [62.7, 65.2] & 0 & 0.0015\\
\midrule
View, full minus view net use & 64.1 & [62.8, 65.3] & 0 & 0.0015\\
View minus full, permitted use & 0.5 & [-0.2, 1.2] & $-5$ & ---\\
View minus event removal, retention & 67.8 & [65.4, 70.1] & 0 & ---\\
\bottomrule
\end{tabular*}\end{lrbox}
\ifdim\wd\OCtable>\linewidth\PackageError{access-paper}{Table tab:confirm exceeds its column}{Revise the table layout.}\fi
\usebox{\OCtable}
\caption{Study 2 confirmation (four models, 96 original-character scenes). The channel test requires all four channels to be positive, the state-correct test uses replies whose state probe is correct (6,068), and the view test requires the view to reduce net use, keep permitted use within 5 pp, and keep more observations than removal. One Holm-adjusted $p$ per family.}
\label{tab:confirm}
\end{table*}
\begin{table*}[!tp]
\centering\setlength{\tabcolsep}{4.5pt}\small
\begin{lrbox}{\OCtable}\begin{tabular*}{\linewidth}{@{\extracolsep{\fill}}lrrrrrr@{}}
\toprule
\textbf{Model} & \textbf{Hear} & \textbf{Understand} & \textbf{See} & \textbf{Receive} & \textbf{State correct} & \textbf{Conditional $D$}\\
\midrule
Gemma-4-31B & 5.5 & 0.0 & 16.4 & 22.4 & 100.0 & 11.1\\
GLM-4.7-Flash & 87.2 & 71.1 & 66.4 & 52.6 & 99.9 & 69.3\\
Nemotron-3-Nano & 99.0 & 66.4 & 97.7 & 99.0 & 95.2 & 90.0\\
Qwen3.5-9B & 97.4 & 51.3 & 96.6 & 96.4 & 100.0 & 85.4\\
\bottomrule
\end{tabular*}\end{lrbox}
\ifdim\wd\OCtable>\linewidth\PackageError{access-paper}{Table tab:modeldetail exceeds its column}{Revise the table layout.}\fi
\usebox{\OCtable}
\caption{Per-model results of the Study 2 confirmation set. Channel columns and conditional $D$ are net use in pp, and state correctness is a percentage.}
\label{tab:modeldetail}
\end{table*}
\begin{table*}[!tp]
\centering\setlength{\tabcolsep}{4.5pt}\small
\begin{lrbox}{\OCtable}\begin{tabular*}{\linewidth}{@{\extracolsep{\fill}}llrrrr@{}}
\toprule
\textbf{Model} & \textbf{Channel} & \textbf{No correct (/192)} & \textbf{Yes correct (/192)} & \textbf{False Yes} & \textbf{False No}\\
\midrule
Gemma-4-31B & Hear & 192 (100.0\%) & 192 (100.0\%) & 0 & 0\\
Gemma-4-31B & Understand & 192 (100.0\%) & 192 (100.0\%) & 0 & 0\\
Gemma-4-31B & See & 192 (100.0\%) & 187 (97.4\%) & 0 & 5\\
Gemma-4-31B & Receive & 192 (100.0\%) & 192 (100.0\%) & 0 & 0\\
GLM-4.7-Flash & Hear & 191 (99.5\%) & 192 (100.0\%) & 1 & 0\\
GLM-4.7-Flash & Understand & 192 (100.0\%) & 192 (100.0\%) & 0 & 0\\
GLM-4.7-Flash & See & 192 (100.0\%) & 178 (92.7\%) & 0 & 14\\
GLM-4.7-Flash & Receive & 192 (100.0\%) & 190 (99.0\%) & 0 & 2\\
Nemotron-3-Nano & Hear & 155 (80.7\%) & 192 (100.0\%) & 37 & 0\\
Nemotron-3-Nano & Understand & 192 (100.0\%) & 192 (100.0\%) & 0 & 0\\
Nemotron-3-Nano & See & 192 (100.0\%) & 141 (73.4\%) & 0 & 51\\
Nemotron-3-Nano & Receive & 192 (100.0\%) & 192 (100.0\%) & 0 & 0\\
Qwen3.5-9B & Hear & 192 (100.0\%) & 192 (100.0\%) & 0 & 0\\
Qwen3.5-9B & Understand & 192 (100.0\%) & 192 (100.0\%) & 0 & 0\\
Qwen3.5-9B & See & 192 (100.0\%) & 175 (91.2\%) & 0 & 17\\
Qwen3.5-9B & Receive & 192 (100.0\%) & 192 (100.0\%) & 0 & 0\\
\bottomrule
\end{tabular*}\end{lrbox}
\ifdim\wd\OCtable>\linewidth\PackageError{access-paper}{Table tab:probecontrols exceeds its column}{Revise the table layout.}\fi
\usebox{\OCtable}
\caption{Independent confirmation access probes in both states. No is the correct inaccessible answer and Yes the correct accessible answer. Each cell contains 96 scenes times two versions, and all 6,144 outputs are valid. False Yes and False No count the two error directions. Model-level accuracy averages four equally sized channels, and always answering No gives 50\% on the balanced 3,072/3,072 population.}
\label{tab:probecontrols}
\end{table*}
\begin{table*}[!tp]
\centering\setlength{\tabcolsep}{4.5pt}\small
\begin{lrbox}{\OCtable}\begin{tabular*}{\linewidth}{@{\extracolsep{\fill}}llrlr@{}}
\toprule
\textbf{Channel} & \textbf{State-correct $D$ [95\% CI]} & \textbf{$N$} & \textbf{Capability-eligible $D$ [95\% CI]} & \textbf{$N$}\\
\midrule
Hear & 72.2 [70.8, 73.5] & 1460 & 72.2 [70.8, 73.5] & 1460\\
Understand & 47.2 [44.9, 49.6] & 1536 & 51.0 [48.4, 53.7] & 1152\\
See & 69.3 [67.3, 71.4] & 1536 & 69.3 [67.3, 71.4] & 1536\\
Receive & 67.6 [65.8, 69.3] & 1536 & 67.6 [65.8, 69.3] & 1536\\
\bottomrule
\end{tabular*}\end{lrbox}
\ifdim\wd\OCtable>\linewidth\PackageError{access-paper}{Table tab:stateboundary exceeds its column}{Revise the table layout.}\fi
\usebox{\OCtable}
\caption{State-correct net use for the four models of Section~\ref{sec:s2} (pp). The capability-eligible column keeps only model and language units whose translation probe reaches 85\%, which changes only comprehension.}
\label{tab:stateboundary}
\end{table*}
\begin{table*}[!tp]
\begin{minipage}[t]{0.47\linewidth}
\centering\setlength{\tabcolsep}{4.5pt}\small
\begin{lrbox}{\OCtable}\begin{tabular}{@{}lrrr@{}}
\toprule
\textbf{Model} & \textbf{Chinese} & \textbf{Japanese} & \textbf{Spanish}\\
\midrule
Gemma-4-31B & 100.0 & 100.0 & 100.0\\
GLM-4.7-Flash & 100.0 & 62.5 & 100.0\\
Nemotron-3-Nano & 100.0 & 75.0 & 87.5\\
Qwen3.5-9B & 100.0 & 62.5 & 100.0\\
\bottomrule
\end{tabular}\end{lrbox}
\adjustbox{max width=\linewidth}{\usebox{\OCtable}}
\caption{Translation probe accuracy (\%), the share of 64 prompts per model and language in which both hidden values are translated. Main estimates use all units, and the capability-eligible estimates of Table~\ref{tab:stateboundary} drop units below 85\%.}
\label{tab:translate}
\end{minipage}\hfill
\begin{minipage}[t]{0.47\linewidth}
\centering\setlength{\tabcolsep}{4.5pt}\small
\begin{lrbox}{\OCtable}\begin{tabular}{@{}lll@{}}
\toprule
\textbf{Channel} & \textbf{Membership minus flag} & \textbf{Prose+flag minus flag}\\
\midrule
Hear & 17.5 [15.0, 20.1] & -37.6 [-41.3, -33.9]\\
Understand & 28.6 [25.5, 31.8] & -4.0 [-8.1, -0.1]\\
See & 32.9 [29.0, 36.8] & -17.5 [-22.5, -13.4]\\
Receive & 3.4 [0.6, 6.3] & -37.9 [-41.4, -34.1]\\
\bottomrule
\end{tabular}\end{lrbox}
\adjustbox{max width=\linewidth}{\usebox{\OCtable}}
\caption{Representation effects on net use (pp), with 95\% intervals. Eleven discovery models, 24 matched scenes. The same comprehension prose contrast is +5.7 [0.0, 12.0] across the four core models.}
\label{tab:representation}
\end{minipage}
\end{table*}

\begin{table*}[!tp]
\centering\setlength{\tabcolsep}{4.5pt}\small
\begin{lrbox}{\OCtable}\begin{tabular*}{\linewidth}{@{\extracolsep{\fill}}lrrr@{}}
\toprule
\textbf{Method} & \textbf{Net use} & \textbf{Permitted use} & \textbf{Event retention}\\
\midrule
Full context & 64.1 & 92.8 & 73.3\\
Event removal & 0.0 & 93.3 & 32.3\\
Character view & 0.0 & 93.3 & 100.0\\
\bottomrule
\end{tabular*}\end{lrbox}
\ifdim\wd\OCtable>\linewidth\PackageError{access-paper}{Table tab:methods exceeds its column}{Revise the table layout.}\fi
\usebox{\OCtable}
\caption{Confirmation set. Net use $D$ is in pp, and permitted use and partial-event retention are percentages. Retention excludes Hear.}
\label{tab:methods}
\end{table*}
\begin{table*}[!tp]
\centering\setlength{\tabcolsep}{4.5pt}\small
\begin{lrbox}{\OCtable}\begin{tabular*}{\linewidth}{@{\extracolsep{\fill}}llll@{}}
\toprule
\textbf{Channel} & \textbf{Method} & \textbf{Inaccessible $D$} & \textbf{Permitted use}\\
\midrule
Hear & Full & 57.2 [54.7, 59.6] & 89.7 [86.6, 92.1]\\
Hear & Prose+flag & 34.4 [31.2, 37.6] & 93.4 [90.7, 95.7]\\
Hear & Self-filter & 47.7 [44.6, 50.9] & 93.5 [91.1, 95.7]\\
Hear & Event removal & 0.1 [0.0, 0.3] & 89.1 [85.4, 92.0]\\
Hear & Character view & 0.1 [0.0, 0.3] & 89.0 [85.3, 92.0]\\
Understand & Full & 31.0 [27.7, 33.9] & 91.5 [88.5, 94.0]\\
Understand & Prose+flag & 32.3 [29.3, 35.1] & 95.6 [93.7, 97.3]\\
Understand & Self-filter & 2.0 [0.9, 3.4] & 62.9 [59.6, 66.1]\\
Understand & Event removal & 0.0 [0.0, 0.0] & 87.8 [84.1, 91.0]\\
Understand & Character view & -0.1 [-0.3, 0.0] & 87.8 [84.2, 91.0]\\
See & Full & 45.7 [41.5, 49.5] & 79.3 [75.2, 83.2]\\
See & Prose+flag & 27.2 [24.6, 29.6] & 96.6 [95.2, 97.7]\\
See & Self-filter & 22.2 [19.8, 24.6] & 81.5 [78.2, 84.5]\\
See & Event removal & 0.0 [0.0, 0.0] & 80.6 [76.4, 84.4]\\
See & Character view & -0.1 [-0.3, 0.0] & 80.5 [76.4, 84.4]\\
Receive & Full & 56.8 [53.7, 59.6] & 94.2 [91.0, 96.6]\\
Receive & Prose+flag & 31.3 [29.2, 33.4] & 96.8 [95.6, 97.8]\\
Receive & Self-filter & 4.8 [3.2, 6.6] & 53.0 [48.9, 57.2]\\
Receive & Event removal & -0.1 [-0.3, 0.0] & 90.9 [87.6, 93.7]\\
Receive & Character view & 0.0 [-0.3, 0.3] & 90.9 [87.6, 93.7]\\
\bottomrule
\end{tabular*}\end{lrbox}
\ifdim\wd\OCtable>\linewidth\PackageError{access-paper}{Table tab:discoverymethods exceeds its column}{Revise the table layout.}\fi
\usebox{\OCtable}
\caption{Five methods on six discovery models and 48 matched scenes (1,152 endpoints per cell). Net use is in pp and permitted use is a percentage. Intervals are 95\% paired-bootstrap intervals.}
\label{tab:discoverymethods}
\end{table*}
\begin{table*}[!tp]
\centering\setlength{\tabcolsep}{4.5pt}\small
\begin{lrbox}{\OCtable}\begin{tabular*}{\linewidth}{@{\extracolsep{\fill}}lllll@{}}
\toprule
\textbf{Channel} & \textbf{Method} & \textbf{Delay 0} & \textbf{Delay 2} & \textbf{Delay 4}\\
\midrule
Hear & full & 69.1 [65.2, 72.7] & 68.8 [65.6, 71.9] & 71.1 [68.0, 73.8]\\
Hear & view & 0.0 [0.0, 0.0] & 0.0 [0.0, 0.0] & 0.0 [0.0, 0.0]\\
Understand & full & 42.6 [36.7, 48.1] & 52.7 [48.1, 57.0] & 58.2 [53.1, 63.3]\\
Understand & view & 0.0 [0.0, 0.0] & 0.0 [0.0, 0.0] & 0.0 [0.0, 0.0]\\
See & full & 55.5 [49.6, 61.3] & 50.0 [44.1, 55.9] & 50.8 [43.8, 57.0]\\
See & view & 0.0 [0.0, 0.0] & 0.0 [0.0, 0.0] & 0.0 [0.0, 0.0]\\
Receive & full & 68.0 [63.7, 71.9] & 68.4 [63.7, 72.7] & 69.5 [66.0, 72.7]\\
Receive & view & -0.4 [-1.2, 0.0] & 0.0 [0.0, 0.0] & 0.0 [0.0, 0.0]\\
\bottomrule
\end{tabular*}\end{lrbox}
\ifdim\wd\OCtable>\linewidth\PackageError{access-paper}{Table tab:delay exceeds its column}{Revise the table layout.}\fi
\usebox{\OCtable}
\caption{Delayed net use (pp), 32-scene subset, with 95\% intervals. The original event remains in context throughout the scripted exchanges.}
\label{tab:delay}
\end{table*}
\begin{table*}[!tp]
\centering\setlength{\tabcolsep}{4.5pt}\small
\begin{lrbox}{\OCtable}\begin{tabular*}{\linewidth}{@{\extracolsep{\fill}}llrrr@{}}
\toprule
\textbf{Channel} & \textbf{Method} & \textbf{Change in choosing A} & \textbf{95\% CI} & \textbf{Valid choices (\%)}\\
\midrule
Hear & full & 63.3 & [53.9, 71.1] & 99.6\\
Hear & view & 0.8 & [0.0, 2.3] & 96.9\\
Understand & full & 54.7 & [47.7, 60.9] & 100.0\\
Understand & view & 0.0 & [0.0, 0.0] & 100.0\\
See & full & 49.2 & [43.0, 54.7] & 100.0\\
See & view & 0.0 & [0.0, 0.0] & 99.2\\
Receive & full & 55.5 & [46.9, 64.1] & 99.6\\
Receive & view & 0.0 & [0.0, 0.0] & 97.7\\
\bottomrule
\end{tabular*}\end{lrbox}
\ifdim\wd\OCtable>\linewidth\PackageError{access-paper}{Table tab:decision exceeds its column}{Revise the table layout.}\fi
\usebox{\OCtable}
\caption{Change in choosing place A (pp) when the hidden fact switches from B to A, and the share of valid choices. Place A is tracked across versions, independently of its option number.}
\label{tab:decision}
\end{table*}
\begin{table*}[!tp]
\centering\setlength{\tabcolsep}{4.5pt}\small
\begin{lrbox}{\OCtable}\begin{tabularx}{\textwidth}{@{}lX@{}}
\toprule
\textbf{Input component} & \textbf{Exact saved text or specified substitution}\\
\midrule
Role system prefix & {\footnotesize\ttfamily You are role-playing Mira Sen. Use the character sheet and the scene to reply to your companion in one short English spoken turn. Return only the spoken words.}\\
Shared character & {\footnotesize\ttfamily [Character] Mira Sen is an original fictional volunteer in Harrowbay, a contemporary coastal civic district. They are practical, polite, and concise. They rely on their own observations. Their history and relations are defined only by this scene.}\\
Full scene (A, no access) & {\footnotesize\ttfamily [Scene state] \{"character\_id": "Mira Sen", "events": [\{"actor": "Alex", "audible\_to\_Mira Sen": false, "boundary\_channel": "Hear", "content": "The glove is in the church.", "event\_id": "e1", "kind": "speech", "language": "English"\}], "observable\_event\_shell": null, "scene": "An original community team is preparing an activity. Its members have not shared the latest finding."\}}\\
Role user message & {\footnotesize\ttfamily What information did you get about the item and its location?}\\
Probe prefix & {\footnotesize\ttfamily Read the scene as an evaluator. Answer the access question with only Yes or No.}\\
Probe question & {\footnotesize\ttfamily Did Mira Sen hear the utterance?}\\
Counterfactual B & \texttt{The glove is in the church.} $\rightarrow$ \texttt{The broom is in the stairwell.}\\
Accessible control & \texttt{audible\_to\_Mira Sen}: \texttt{false} $\rightarrow$ \texttt{true}\\
Inaccessible view scene & {\footnotesize\ttfamily [Scene state] \{"character\_id": "Mira Sen", "observations": [], "scene": "An original community team is preparing an activity. Its members have not shared the latest finding."\}}\\
\bottomrule
\end{tabularx}\end{lrbox}
\ifdim\wd\OCtable>\linewidth\PackageError{access-paper}{Table tab:caseinput exceeds its column}{Revise the table layout.}\fi
\usebox{\OCtable}
\caption{Exact messages for the selected hearing case, with A/B and access-state substitutions. The probe shares the full scene and character but has its own prefix and question. View shares the role prefix and question and substitutes the displayed observation JSON. Accessible view contains the permitted event in its observation list.}
\label{tab:caseinput}
\end{table*}
\begin{table*}[!tp]
\centering\setlength{\tabcolsep}{4.5pt}\small
\begin{lrbox}{\OCtable}\begin{tabularx}{\textwidth}{@{}llcXX@{}}
\toprule
\textbf{Twin} & \textbf{Access} & \textbf{Probe} & \textbf{Full-context spoken reply} & \textbf{Character-view spoken reply}\\
\midrule
A & No & No & I heard Alex mention the glove is in the church, though he wasn't speaking to me directly. & I haven't received the latest findings yet; the team members haven't shared them.\\
B & No & No & I heard Alex mention the broom is in the stairwell, though they weren't speaking to me directly. & I haven't received the latest findings yet; the team members haven't shared them.\\
A & Yes & Yes & Alex mentioned the glove is in the church. & Alex mentioned the glove is in the church.\\
B & Yes & Yes & Alex mentioned the broom is in the stairwell. & Alex just told me the broom is in the stairwell.\\
\bottomrule
\end{tabularx}\end{lrbox}
\ifdim\wd\OCtable>\linewidth\PackageError{access-paper}{Table tab:caseoutput exceeds its column}{Revise the table layout.}\fi
\usebox{\OCtable}
\caption{All saved content replies and independent probe answers for scene C001, Qwen3.5-9B, access flag, no delay. All four full-context replies name the target and not the foil ($D=1$). Under the character view, the inaccessible replies name neither ($D=0$) and the accessible replies name the target ($D=1$).}
\label{tab:caseoutput}
\end{table*}
\begin{table*}[!tp]
\centering\setlength{\tabcolsep}{4.5pt}\small
\begin{lrbox}{\OCtable}\begin{tabular*}{\linewidth}{@{\extracolsep{\fill}}lllll@{}}
\toprule
\textbf{Channel} & \textbf{Method} & \textbf{A first (18 scenes)} & \textbf{A second (14 scenes)} & \textbf{Equal-position effect}\\
\midrule
Hear & full & 65.3 [56.9, 72.2] & 60.7 [42.9, 75.0] & 63.0 [53.3, 71.2]\\
Hear & view & 0.0 [0.0, 0.0] & 1.8 [0.0, 5.4] & 0.9 [0.0, 2.7]\\
Understand & full & 56.9 [50.0, 63.9] & 51.8 [39.3, 62.5] & 54.4 [47.1, 60.9]\\
Understand & view & 0.0 [0.0, 0.0] & 0.0 [0.0, 0.0] & 0.0 [0.0, 0.0]\\
See & full & 50.0 [43.1, 56.9] & 48.2 [39.3, 57.1] & 49.1 [42.8, 54.9]\\
See & view & 0.0 [0.0, 0.0] & 0.0 [0.0, 0.0] & 0.0 [0.0, 0.0]\\
Receive & full & 59.7 [50.0, 69.4] & 50.0 [35.7, 62.5] & 54.9 [45.5, 63.1]\\
Receive & view & 0.0 [0.0, 0.0] & 0.0 [0.0, 0.0] & 0.0 [0.0, 0.0]\\
\bottomrule
\end{tabular*}\end{lrbox}
\ifdim\wd\OCtable>\linewidth\PackageError{access-paper}{Table tab:choiceorder exceeds its column}{Revise the table layout.}\fi
\usebox{\OCtable}
\caption{Option-position sensitivity (pp, 95\% intervals). A/B order and the question are identical within every pair. The equal-position estimate weights the two order strata equally and resamples scenes independently within each stratum.}
\label{tab:choiceorder}
\end{table*}
\begin{table*}[!tp]
\centering\sffamily\footnotesize\renewcommand{\arraystretch}{1.12}\setlength{\tabcolsep}{3.2pt}
\begin{tabular*}{\linewidth}{@{\extracolsep{\fill}}lrrrrrr@{}}
\toprule
\textbf{Model} & \textbf{Base} & \textbf{Sys.} & \textbf{Sheet} & \textbf{State} & \textbf{Quest.} & \textbf{All}\\
\midrule
Claude Sonnet 5.5 & 0.5 & \textbf{9.9} & \textbf{15.1} & \textbf{8.9} & \textbf{9.4} & \textbf{13.0}\\
Gemini 3.1 Pro & 0.0 & -- & 0.0 & 0.0 & -- & \textbf{21.4}\\
\bottomrule
\end{tabular*}
\caption{Single-feature swaps for two proprietary models on the discovery hearing prompts (192 unheard prompts per cell, net use in pp). Bold marks a change from Base whose paired 95\% interval excludes zero. Gemini 3.1 Pro was run on the two swaps that moved the open-weight models most.}
\label{tab:hostedswaps}
\end{table*}
\begin{table*}[!tp]
\centering\setlength{\tabcolsep}{4.5pt}\small
\begin{tabular*}{\linewidth}{@{\extracolsep{\fill}}lrrrrr@{}}
\toprule
\textbf{Model} & \textbf{Study 1 prompt} & \textbf{+ channel} & \textbf{+ question} & \textbf{+ minimal} & \textbf{All three}\\
\midrule
\multicolumn{6}{@{}l}{\emph{Not-audible flag}}\\
Gemma-4-31B & 52.1 & 52.1 & 51.0 & 54.7 & 50.0\\
Qwen3.8-27B & 71.4 & 74.0 & 80.2 & 51.6 & 32.8\\
Seed-OSS-36B & 90.1 & 88.0 & 100.0 & 87.0 & 83.3\\
Phi-4 & 99.0 & 100.0 & 94.3 & 100.0 & 94.3\\
\multicolumn{6}{@{}l}{\emph{Not-understood flag}}\\
Gemma-4-31B & 6.2 & 3.1 & 1.6 & 3.1 & 0.0\\
Qwen3.8-27B & 35.9 & 34.4 & 39.6 & 12.0 & 14.1\\
Seed-OSS-36B & 69.8 & 67.2 & 95.3 & 37.0 & 31.8\\
Phi-4 & 88.5 & 89.6 & 66.1 & 69.8 & 49.5\\
\bottomrule
\end{tabular*}
\caption{Format ablation from Study 1 toward Study 2 (96 scenes, one language per skeleton, content question). Net use (pp) when one Study 2 feature is added (a field naming the channel, the direct question, a minimal context) and with all three.}
\label{tab:ablation}
\end{table*}

\end{document}